\documentclass{article} 
\usepackage{iclr2025_conference,times}

\usepackage{amsmath,amsfonts,bm}

\def\eqref#1{equation~\ref{#1}}

\def\1{\bm{1}}

\DeclareMathAlphabet{\mathsfit}{\encodingdefault}{\sfdefault}{m}{sl}
\SetMathAlphabet{\mathsfit}{bold}{\encodingdefault}{\sfdefault}{bx}{n}

\usepackage{hyperref}
\usepackage{url}
\usepackage[utf8]{inputenc} 
\usepackage[T1]{fontenc}    
\usepackage{hyperref}       
\usepackage{url}            
\usepackage{booktabs}       
\usepackage{amsfonts}       
\usepackage{nicefrac}       
\usepackage{microtype}      
\usepackage{xcolor}         
\usepackage{graphicx}
\usepackage{subcaption}
\usepackage{float}

\title{Preserving Product Fidelity in Large Scale Image Recontextualization with Diffusion Models}

\author{
Ishaan Malhi \thanks{Correspondence to imalhi@google.com, praneetdutta@google.com} \thanks{Google DeepMind} \\
\And
Praneet Dutta \footnotemark[2] \\
\And
Ellie Talius \footnotemark[2] \\
\And
Sally Ma \footnotemark[2] \\
\And
Brendan Driscoll \thanks{Google} \\
\And
Krista Holden \footnotemark[3] \\
\And
Garima Pruthi \footnotemark[3] \\
\And
Arunachalam Narayanaswamy \footnotemark[2] \\
}

\iclrfinalcopy

\begin{document}

\maketitle

\begin{figure}[H]
\centering
\includegraphics[width=0.9\textwidth]{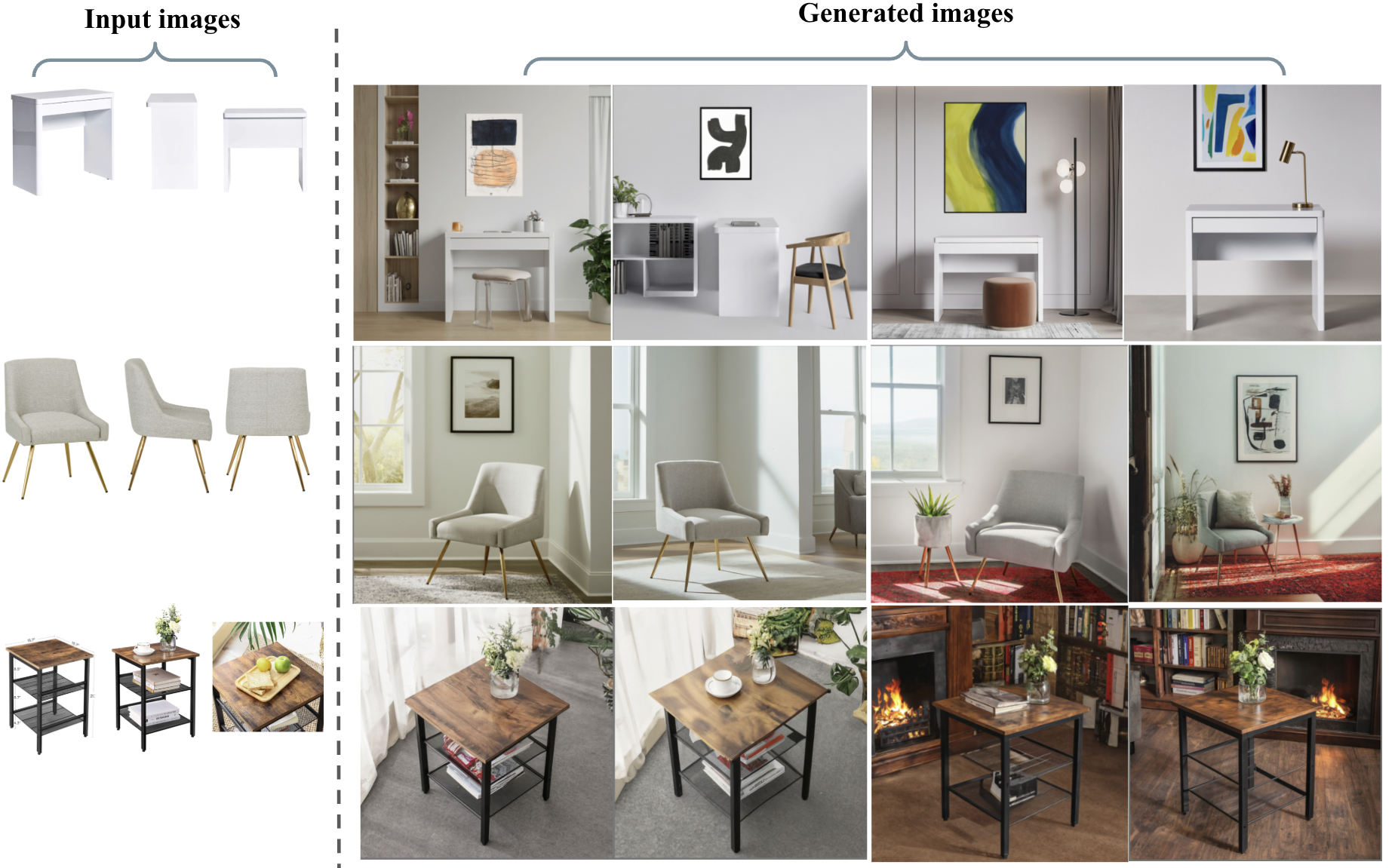}
\caption{Given a few input images of a real world product, our system can generate images that not only maintain high fidelity to the original product, but also recontextualize it in novel settings beyond background changes: from showcasing it in a new perspective, adding object occlusions, to creating different and realistic lighting conditions.}
\label{poster_imgs}
\end{figure}

\begin{abstract}
We present a framework for high-fidelity product image recontextualization using text-to-image diffusion models and a novel data augmentation pipeline. This pipeline leverages image-to-video diffusion, in/outpainting \& negatives to create synthetic training data, addressing limitations of real-world data collection for this task.  Our method improves the quality and diversity of generated images by disentangling product representations and enhancing the model's understanding of product characteristics. Evaluation on the ABO dataset and a private product dataset, using automated metrics and human assessment, demonstrates the effectiveness of our framework in generating realistic and compelling product visualizations, with implications for applications such as e-commerce and virtual product showcasing.
\end{abstract}

\section{Introduction}
Placing products seamlessly into new environments in images holds immense value for e-commerce, VR/AR, content creation among others. Photorealistic recontextualization demands both visual appeal and, crucially, accurate product detail preservation. The most common method that preserves product details, namely background swapping, does not provide realistic product relighting, novel perspectives, or natural occlusions. These shortcomings heavily limit how natural-looking and creative the resulting images can be. While recent advancements in text-to-image diffusion models such as DreamBooth~\citep{ruiz2023dreambooth} and InstructPix2Pix~\citep{brooks2023instructpix2pix} offer promise, directly applying them to diverse product images often compromises fidelity. We observe that these existing methods struggle with intricate details, consistent appearance, and product background disentanglement, especially when dealing with complex textures, reflections, and occlusions in the product images.
Scaling to a wide variety of products amplifies these failure modes, motivating the need for a new recontextualization approach.

We present a novel framework specifically designed to address these limitations and achieve high-fidelity product recontextualization across a wide variety of products. Compared with background replacement and other methods, our system not only preserves product details at scale, but also enables the three capabilities key to blending products seamlessly into new settings: 1) realistic relighting; 2) object occlusions; 3) novel product viewpoints.

\section{Method}
Existing methods \citep{ruiz2023dreambooth} \citep{gal2022imageworthwordpersonalizing} that perform well on subject driven personalization commonly use few shot samples of the product. While simply collecting more samples would improve product fidelity as shown in \citep{ruiz2023dreambooth}, it's not feasible to do so at scale. Instead, we propose augmenting the existing few shot samples through synthetic augmentation to address common failure modes, and finetune the model using LoRA \citep{hu2021loralowrankadaptationlarge}, in a similar setup to DreamBooth \citep{ruiz2023dreambooth}. This method provides simplicity and scalability for a large variety of products.

\subsection{Synthetic Data Augmentation}

Our method addresses the limitations of existing methods by augmenting the training data with synthetically generated examples. We employ a data pipeline consisting of novel view generation, background/object disentanglement, and the inclusion of negative examples (Visualization of the pipeline in Appendix \ref{subset:finetuning_pipeline}). Each stage targets specific failure modes observed in existing methods.

\subsubsection{Novel View Generation}
Increasing the number of input images improves fidelity \citep{ruiz2023dreambooth}, but simple repetition leads to overfitting. We generate novel viewpoints using image-to-video diffusion models \citep{bar2024lumiere}, augmenting the training data with diverse product perspectives (Appendix \ref{synth_app} Table \ref{fig:lumiere_results}). While 3D reconstruction models \citep{barron2023zip, hong2023lrm} offer an alternative, we prioritize video frame interpolation for better prompt adherence and product fidelity. Figure \ref{fig:lumiere_results} in Appendix  \ref{synth_app} demonstrates an example.

\subsubsection{New Context Images}
To reduce overfitting to the background elements in input images, we  used masked outpainting to change the backgrounds of the base, and novel view images. These images serve as positives while helping disentangle the product from the backgrounds during finetuning. We used an LLM to generate new prompts for each image, then created a segmentation mask using \citep{kirillov2023segment}. The mask and new context prompt was used to outpaint the image to change the background. To reduce hallucinations we curate and cache the context prompts. The caching strategy is described in Appendix subsection~\ref{subsec:prompt_bank}.

\subsubsection{Augment with Negatives}
While \citep{ruiz2023dreambooth} uses a class prior preservation loss, we observed that adding negative images has a similar effect. We additionally added counterfactuals - images with the same background as the ground truth images, but with different objects inpainted into the product's masked region. We observed this gave the following benefits: (1) preserve class priors in the base generation model, (2) accurately render non-product objects in the background/foreground (3) render products with reduced artifacts in the image. To generate a negative image we took the caption of a fine tuning image,  and generated a new image using the base diffusion model. We observe that a 2:1 positive:negative image ratio produced higher image quality and fidelity at the cost of diversity.

\subsection{Captioning}
We used an internally trained captioning model based on a Vision Language Model (VLM) \citep{geminiteam2024geminifamilyhighlycapable} that is provided the base set of images with an instruction asking for fine grained details of the image, including various attributes like color, position, lighting, and other fine grained product details. We found that fine grained captioning greatly improves subject recontextualization quality.

\subsection{Training Data Filtering}
While generating data using the approaches mentioned above greatly increases the amount of samples, some images may be higher quality than others. For example, it is possible for outpainting to produce hallucinations that extend the products or add new attributes. We developed techniques to reduce errors from outpainting in two ways to improve the quality of our finetuning training set. Particularly, 1) Use automated metrics to evaluate each additional outpainted image and sample the highest rated images. Specifically, we use CLIP \citep{2021clip} (both CLIP-I and CLIP-T) as well as segmented CLIP Embedding for metrics; and 2) Compare the segmentation mask of the outpainted image and the reference image, and filter out outpainted images whose IoUs (Intersection over Unions) for the two masks fall below an experimentally determined threshold.

\subsection{Model Finetuning}
We chose to adopt the approach described in \citep{ruiz2023dreambooth}, with three adjustments: Class prior preservation loss is omitted in favor of more negatives during finetuning, we trained on a higher learning rate and for longer and we swept over possible synthetic tokens to be used for each product. We observed that some tokens perform significantly better than the others. Tokens that can be confused as product names are avoided and we chose to sample from an existing list of rare tokens that have plausible association with the product being finetuned. We find that using LoRA finetuning produced better results compared to finetuning the whole model.

\subsection{Post Finetuning Ranking}
We re-use the evaluation metrics for images to rank and pick the top N images above an experimentally determined threshold. We use CLIP-T, CLIP-I, DINO \citep{caron2021emergingpropertiesselfsupervisedvision} embeddings to rank images on product fidelity. We pick the top N images per product after ranking them by the aggregate sum of these metrics. This leads to a relative 30.9\% increase in image pass rate. Our analysis also showed a 0.4 Pearson correlation between these automatic metrics and human ratings. As a future direction, a ranking/classification model can be learned to maximize correlation with human ratings.

\section{Results}
We evaluated our proposed framework on the Amazon Berkeley Objects (ABO) \citep{collins2022abo} dataset, creating a challenging dataset due to its diverse range of product categories and complex background scenes. We also used a private set of consumer products, similar to the ABO dataset in it's difficulty and range of product types. We quantitatively assess the fidelity of our generated images using established metrics, including CLIP and DINO based image similarity scores. We also conduct a qualitative analysis through human evaluation, comparing our approach to an existing baseline of DreamBooth + LoRA.

\subsection{Qualitative results}
We showcase results on products in the ABO dataset in Figure \ref{ABO_results}. Our method preserves the product details with high fidelity while seamlessly integrating the product into a realistic lifestyle scene. We observe that the generated images exhibit diverse and plausible arrangements of furniture and decor, consistent with the provided prompt. Our method is able to achieve the following (1) object occlusions, (2) novel view generation and (3) varied realistic lighting conditions, particularly foreground relighting. In Figure \ref{ABO_results} we demonstrate the high quality scene composition of the method, and exemplify its advantage over background replacement systems, which do not exhibit these features. We also showcase additional qualitative results from ablation studies focusing on LoRA rank and training steps in section \ref{Lora Ablation Section}

\subsection{Quantitative Results}
We provide human evaluation results as well as automated metrics on our approach. For evaluation, we selected a closed set of real world objects distinct from the ABO dataset. A small closed set of 100 products were used to account for the cost of human evaluation, reduce any effects of data contamination and demonstrate the generality of our approach to other object datasets. We compare our method to a baseline of using just the method from \citep{ruiz2023dreambooth} and LoRA.

\subsubsection{Human Evaluation Results}
For human evaluation, we showed raters the source and generated images and asked them to accept or reject the generated images. A series of 8 questions all needed to pass and we took a majority vote over 3 raters (Additional details in Appendix \ref{subsec:human_eval_guidelines}). We report the pass rate per image, and per product (defined as the \% of products with at least one passing image). Human evaluations show that our method performs significantly better than a baseline method of DreamBooth + LoRA (Rank=64) on a closed evaluation set of various furniture products. Results are reported in Table \ref{tab:human_eval}.

\begin{table}[htbp]
\begin{tabular}{p{0.3\textwidth}p{0.3\textwidth}p{0.3\textwidth}}
\toprule
Dataset & Overall Per Image pass rate $\uparrow$ & Per Product pass rate $\uparrow$\\
\midrule
Furniture (Ours) & \textbf{17.40\%} & \textbf{45.5\%} \\ 
Furniture (DreamBooth) & 10.00\% & 24.00\% \\ 
\bottomrule
\\
\end{tabular}
\caption{Human Evaluation results on multiple datasets created by mixing several product types in a closed evaluation set. Due to variations in generation, we see that per product pass rate is often much higher than the per image pass rate, demonstrating the difficulty of the task.}
\label{tab:human_eval}
\end{table}

\subsubsection{Automatic Evaluation Metrics}
Due to the cost of human evaluation, we also evaluated our method using automatically computed metrics (Table \ref{non-seg-results}). Following ~\citep{ruiz2023dreambooth} we use the CLIP-I and DINO metrics to compare generated images against the reference images to evaluate product fidelity. We also used CLIP-T to evaluate text alignment as from \citep{ruiz2023dreambooth}. We also used SegCLIP-I (and Segmented DINO) for more localized cosine similarity as suggested in \citep{zhu2024multiboothgeneratingconceptsimage}. We achieved consistent performance on metrics on the ABO dataset and our private evaluation set, demonstrating the ability to scale to a wide variety of products.

\begin{table}[H]
\centering
\begin{tabular}{lccc}
\toprule
& ABO & Private Set (Ours) & Private Set (DreamBooth + LoRA Baseline) \\
\midrule
CLIP - I $\uparrow$ & 0.89 & 0.80 & \textbf{0.81} \\
CLIP - T $\uparrow$ & 0.24 & 0.20 & \textbf{0.25} \\
DINO - I $\uparrow$ & 0.98 & \textbf{0.94} & 0.92 \\
Seg CLIP - I $\uparrow$ & 0.94 & \textbf{0.90} & 0.88 \\
Seg CLIP - T $\uparrow$ & 0.21 & \textbf{0.24} & 0.22 \\
Seg DINO - I $\uparrow$ & 0.99 & \textbf{0.99} & 0.98 \\
\bottomrule
\end{tabular}

\caption{Image and Text alignment metrics for our process on ABO dataset and a private evaluation set. The segmented metrics are higher than the non-segmented versions, indicating that the products in reference and generated images are similar, while the background of the images varies.}
\label{non-seg-results}
\label{seg-results}
\end{table}

\section{Conclusion}
We demonstrate an improved data pipeline and an alternate strategy to subject driven generation, particularly for product recontextualization, to millions of products at scale. Experimental observations and human evaluation show that our method performs significantly better than existing approaches while requiring no model surgery and expensive finetuning.

\bibliography{references}
\bibliographystyle{plainnat}

\newpage
\appendix

\section{Appendix / supplemental material}

\begin{table}
\begin{tabular}{lllll}

\end{tabular}
\end{table}

\subsection{Synthetic Augmentation Techniques}
\label{synth_app}

\begin{figure}[htbp]
\centering
\includegraphics[width=0.2\textwidth]{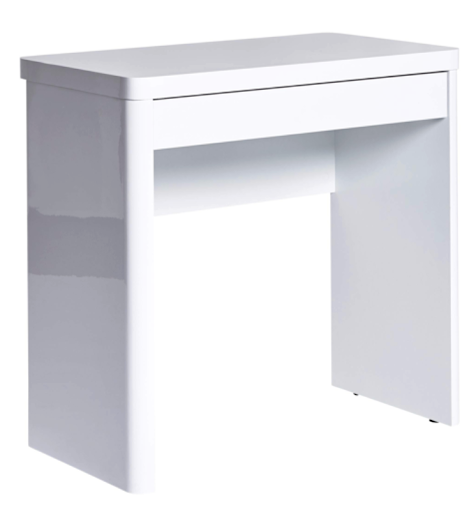}
\includegraphics[width=0.2\textwidth]{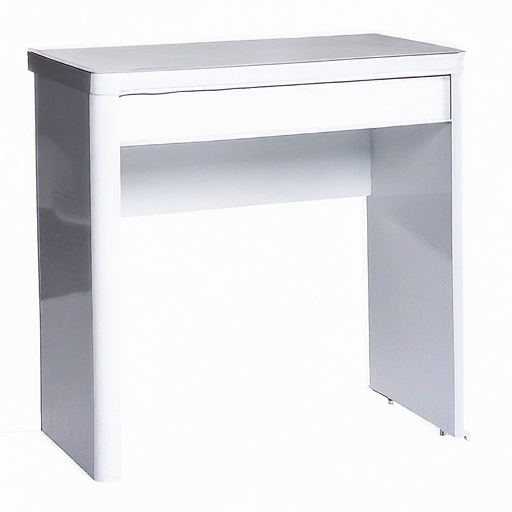}
\caption{Novel view generated using image-to-video diffusion. Left: Input image. Right: Generated view.}
\label{fig:lumiere_results}
\end{figure}

\begin{figure}[h]
\centering
\includegraphics[width=0.3\textwidth]{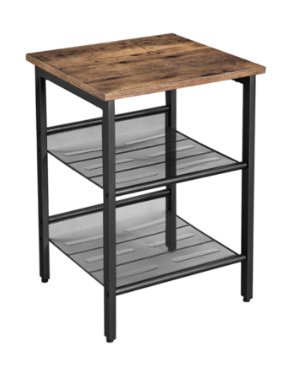}
\includegraphics[width=0.3\textwidth]{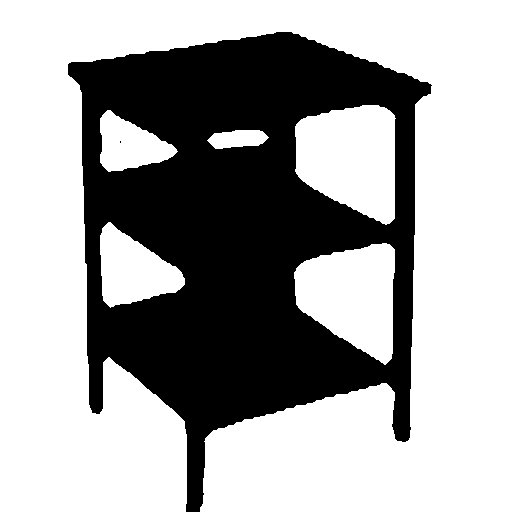}
\includegraphics[width=0.3\textwidth]{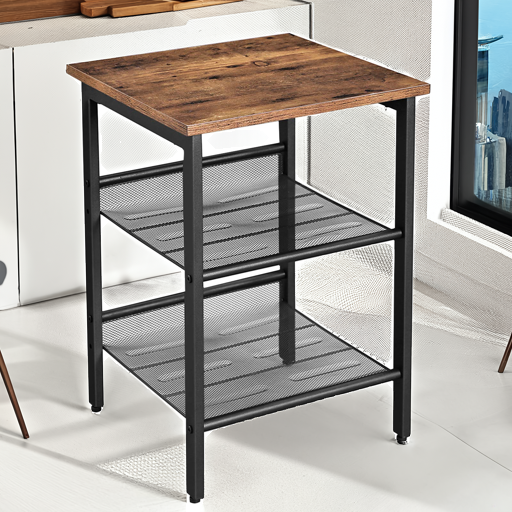}
\caption{Samples of image positives for a given table, along with it's object mask used for background replacement/outpainting.}
\label{fig:positives}
\end{figure}

\begin{figure}[htbp]
\centering
\begin{subfigure}{\textwidth}
\centering
\includegraphics[width=0.3\linewidth]{table_b_1_original.png}
\includegraphics[width=0.3\linewidth]{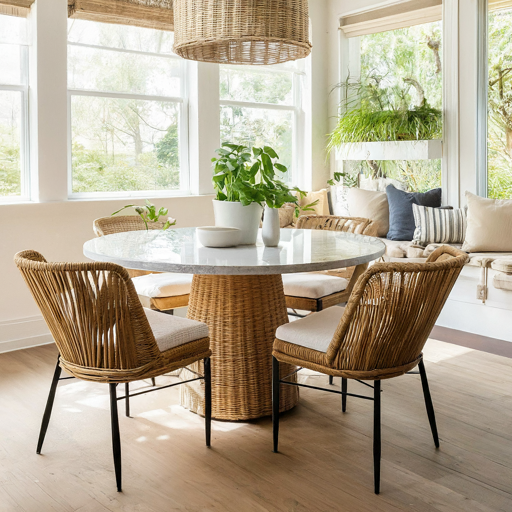}
\includegraphics[width=0.3\linewidth]{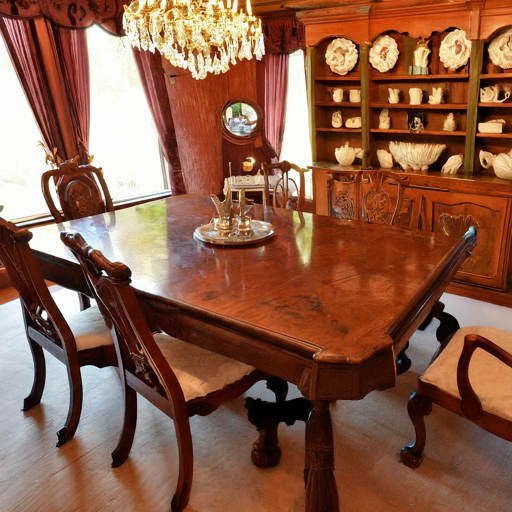}
\caption{Samples of image negatives for a dining table. Left most is the original unmodified image, while the rest are in-class negatives. The prompts used are described in Appendix \ref{subsec:prompts}}
\label{fig:negatives}
\end{subfigure}
\hfill
\begin{subfigure}{\textwidth}
\centering
\includegraphics[width=\textwidth,keepaspectratio]{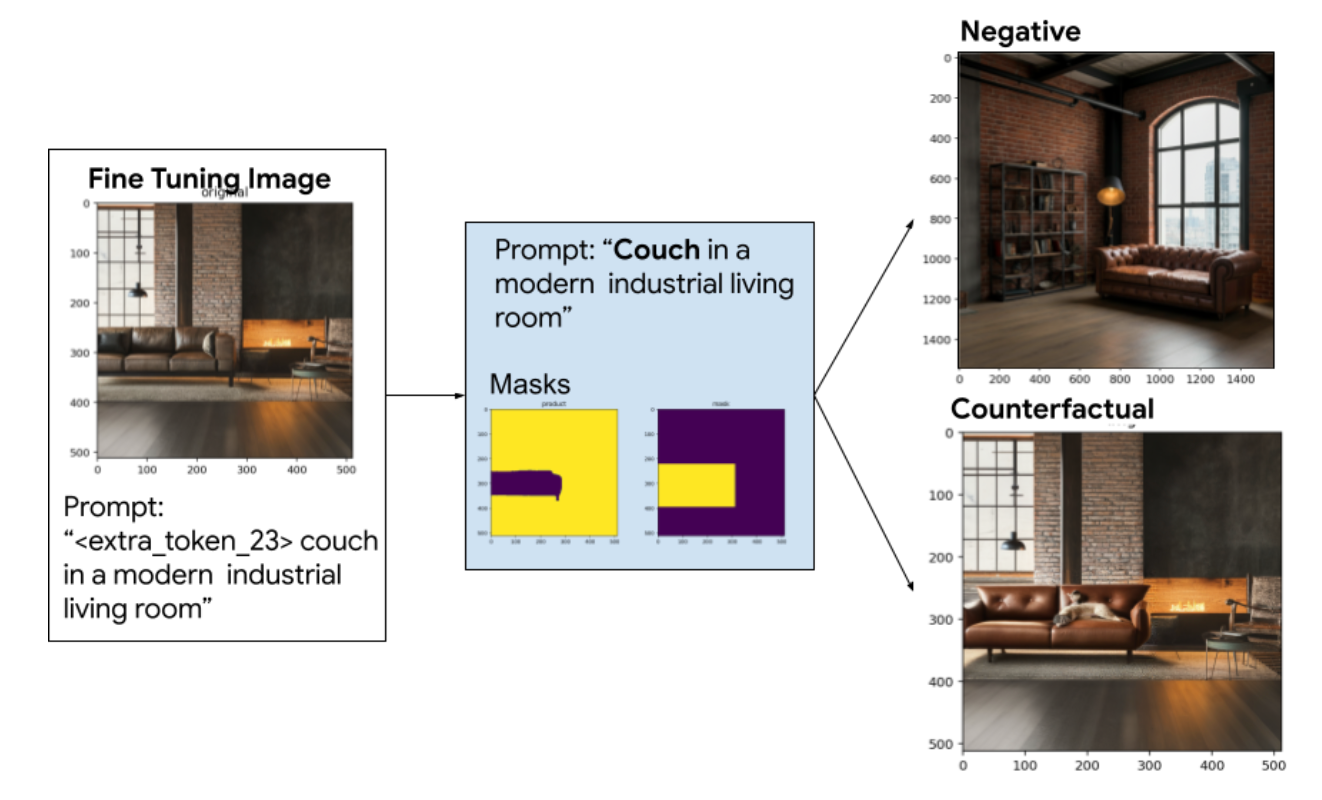}
\caption[\linewidth]{The process for creating counterfactuals and negatives from a starting sample.}
\end{subfigure}
\caption{Samples of counterfactuals and negatives used to augment product images.}
\hfill
\end{figure}
\newpage

\subsection{System Design Choices}
\subsubsection{Finetuning Data Pipeline}
\label{subset:finetuning_pipeline}

\begin{figure}[htbp]
\centering
\includegraphics[width=0.75\textwidth]{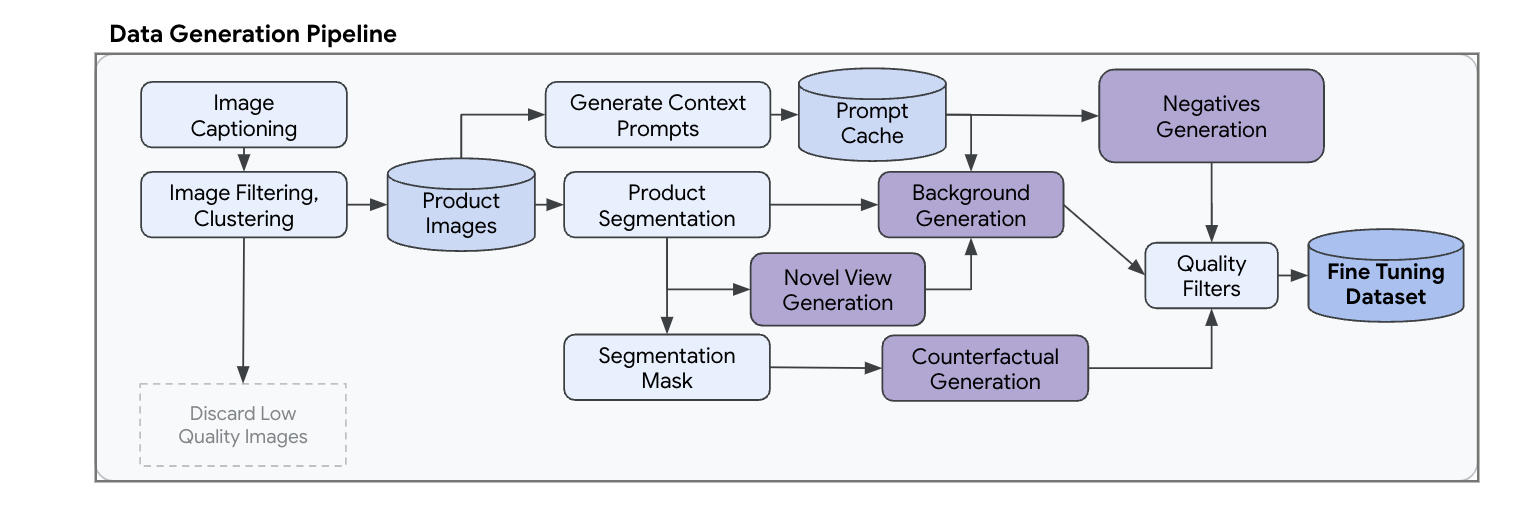}
\caption{Figure illustrating our scalable finetuning data generation approach.}
\end{figure}

\subsubsection{Caching Context Prompts}
\label{subsec:prompt_bank}

Creating context prompts to generate images on the fly, and at scale can be prohibitively expensive and difficult to control for quality. Moreover, while outpainting resulted in novel background images for the same product, generating new prompts for each image was prone to hallucinations. To address this issue, we cached context prompts before running the data pipeline and curated the prompts to remove unrealistic prompts.

To solve these issues, we instead do the following: for each type of product (chairs, phone cases, etc), we generate many possible context prompts using an LLM, and save these prompts in an offline file. We refer to this as a cached "prompt bank". For each individual product, we classify the type of product (either as part of product details available in the dataset metadata or using a VLM), and retrieve prompts from the prompt bank. We then use these prompts for synthetic data augmentations (e.g outpainting) and generating target generation prompts. Using the prompt bank allows us to save compute and improve our prompts over time, while still producing novel contexts for the product.

\newpage

\subsection{Human evaluation guidelines}
\label{subsec:human_eval_guidelines}
Human evaluation guidelines focused on a fine grained evaluation across multiple categories and failure modes. Particularly, 8 questions were posed to human evaluators on a 4 point scale (yes, maybe, no, unclear) - focusing on product fidelity, logo generation fidelity, realistic product use, product size, presence of background/foreground hallucinations, product placement, image safety and likelihood to use the image as a business owner. A 'yes' was considered a pass, and a majority vote over 3 human raters was taken to consider an image to pass human rating.

\subsection{Generated samples}
\label{subsec:prompts}

\begin{figure}[htbp]
\centering
\includegraphics[width=\textwidth,keepaspectratio]{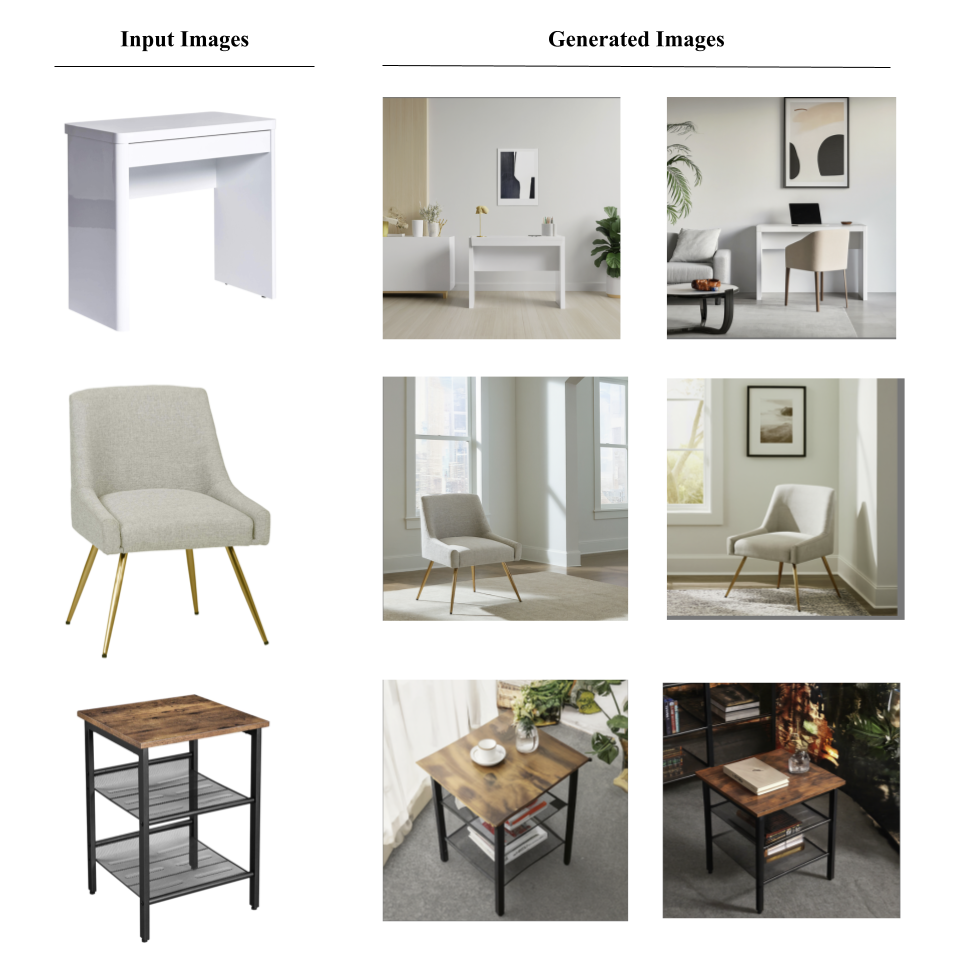}
\caption{Recontextualization results on the ABO dataset, demonstrating high fidelity, object occlusions, novel perspectives, and realistic lighting. Object occlusions are present for the desk and table and all the products demonstrate novel view synthesis, where the angle the object is placed at in the generated images is different than the provided image. The chair highlights the ability of the model to apply realistic and diverse lighting on an object.}
\label{ABO_results}
\end{figure}

\begin{figure}[htbp]
\centering
\begin{subfigure}{0.45\textwidth}
\centering
\includegraphics[width=\linewidth]{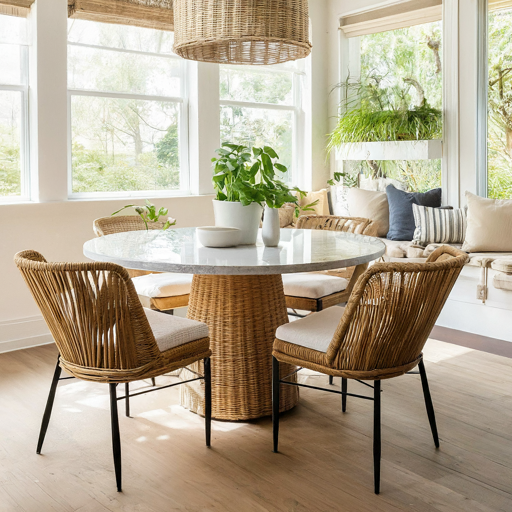}
\caption[\linewidth]{"A round table with a marble top, placed in a sun-drenched breakfast nook with wicker chairs and potted herbs on the windowsill."}
\end{subfigure}
\hfill
\begin{subfigure}{0.45\textwidth}
\centering
\includegraphics[width=\linewidth]{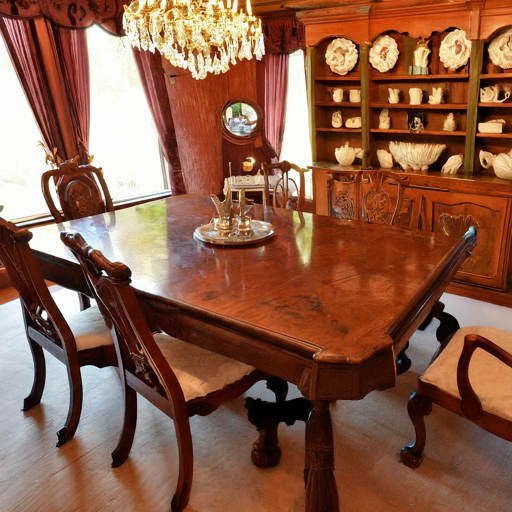}
\caption[\linewidth]{"An antique table with ornate carvings and claw feet, displayed in a grand dining room with a crystal chandelier and antique china on display."}
\end{subfigure}
\hfill
\caption{Prompts used to generate model negative samples.}
\end{figure}

\newpage
\subsection{LoRA Rank Ablation Qualitative Results}
\label{Lora Ablation Section}
\begin{figure}[htbp]
\centering
\includegraphics[width=\textwidth]{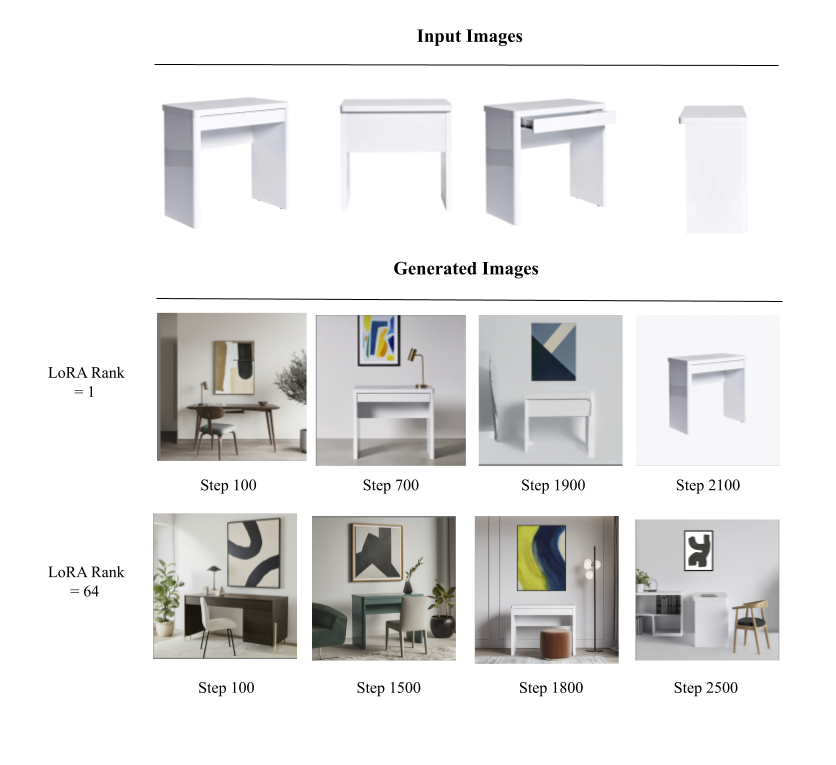}
\caption{Recontextualization results for a desk with varied LoRA rank values and number of steps the model is trained for. These results showcase the tradeoff between object fidelity and generating novel and realistic scenes containing the products. LoRA rank 1 achieves high product fidelity in much fewer training steps but generates less diverse scenes, before eventually overfitting and beginning to reproduce training data regardless of prompt. LoRA rank 64 takes much longer to learn the product, but generates more varied scenes utilizing less common perspectives of the object.  }
\label{Lora_results}
\end{figure}

In addition to the results above, we also showcase qualitative results from ablations on the LoRA rank used during fine tuning and number of training steps in Figure \ref{Lora_results}. We ran our process for the same set of input images of the desk and then finetune a model using LoRA rank=1 and rank=64. As shown in Figure \ref{Lora_results}, as both models were trained for more steps, they improved in the ability to generate an accurate depiction of the desk, and they both lost some amount of complexity and diversity in the scenes they generate.

The rank 1 model learned the appearance of the desk much quicker than rank 64 model (the samples display high product fidelity at 700 steps as opposed to 1800 steps \ref{Lora_results}). When trained for the same number of steps as the rank 64 model, the rank 1 model showed clear signs of overfitting, in some cases even recreating the training data regardless of the prompt given to the model. The rank 1 model also began to generate very simple scenes in step 700, even once it gained the ability to faithfully render the desk.

In contrast, the rank 64 model took much longer to learn the specifics of the desk's appearance, but once it did, we note that the model maintains the ability to generate diverse and realistic scenes much better than the rank 1 model. The scenes from the rank 64 model show the desk in more unique viewpoints and in more complex environments, suggesting the rank 1 model lost this ability while it quickly learned the appearance of the desk. However, even the rank 64 model began to generate less diverse and more simple scenes over time as it learns the appearance of the desk. 

This ablation demonstrates the trade off between product fidelity and generating realistic and diverse scenes. As such, choosing the optimal rank and training steps is equivalent to finding the balance point between product fidelity and generating diverse images. Using a higher LoRA rank allows the model to learn more slowly and better maintain its ability to generate complex and diverse scenes, making it easier to find the balance point for this tradeoff. 

\end{document}